\documentclass[letterpaper, 10 pt, conference]{ieeeconf}  

\IEEEoverridecommandlockouts                              

\usepackage{graphics} 
\usepackage{graphicx}
\usepackage{amsmath} 
\usepackage{amssymb}  
\usepackage{multirow}
\usepackage{soul}  

\usepackage{subcaption}
\title{\LARGE \bf
Rendering Anywhere You See: Renderability Field-guided Gaussian Splatting
}

\author{Xiaofeng Jin$^{*1}$, Yan Fang$^{2}$, Matteo Frosi$^{1}$, Jianfei Ge$^{2}$, Jiangjian Xiao$^{2}$, Matteo Matteucci$^{1}$%
\thanks{This work was conducted at the Laboratory of Artificial Intelligence Robotics and supported by Ningbo Institute of Materials Technology and Engineering, Chinese Academy of Sciences.}
\thanks{*Corresponding author.}
\thanks{$^{1}$Authors are affiliated with Politecnico di Milano, Milan 20133, Italy. {\tt\small \{xiaofeng.jin, matteo.frosi, matteo.matteucci\}@polimi.it}}%
\thanks{$^{2}$Author are affiliated with Ningbo Institute of Materials Technology and Engineering, Chinese Academy of Sciences. {\tt\small \{fangyan, gejianfei, xiaojiangjian\}@nimte.ac.cn}}%
}

\begin{document}

\maketitle
\thispagestyle{empty}
\pagestyle{empty}

\begin{abstract}
Scene view synthesis, which generates novel views from limited perspectives, is increasingly vital for applications like virtual reality, augmented reality, and robotics. Unlike object-based tasks, such as generating 360° views of a car, scene view synthesis handles entire environments where non-uniform observations pose unique challenges for stable rendering quality. To address this issue, we propose a novel approach: renderability field-guided gaussian splatting (RF-GS). This method quantifies input inhomogeneity through a renderability field, guiding pseudo-view sampling to enhanced visual consistency. To ensure the quality of wide-baseline pseudo-views, we train an image restoration model to map point projections to visible-light styles. Additionally, our validated hybrid data optimization strategy effectively fuses information of pseudo-view angles and source view textures. Comparative experiments on simulated and real-world data show that our method outperforms existing approaches in rendering stability.

\end{abstract}

\section{Introduction}\label{sec:intro}
Rendering 3D scenes is crucial for virtual reality/mixed reality (VR/MR) applications~\cite{22}. While current research~\cite{1,2,7,8} achieves high-fidelity rendering, the reliance on dense views limits their practical applicability~\cite{28,27}. Neural Radiance Fields (NeRF)~\cite{1} advanced object reconstruction through neural radiation field representations. While subsequent studies improved its generalization ~\cite{26,29,30}, NeRF's training and rendering costs~\cite{34} remain a limitation.

\begin{figure}[t]
        \centering
	\begin{minipage}[c]{0.47\textwidth}
		\includegraphics[width=\textwidth]{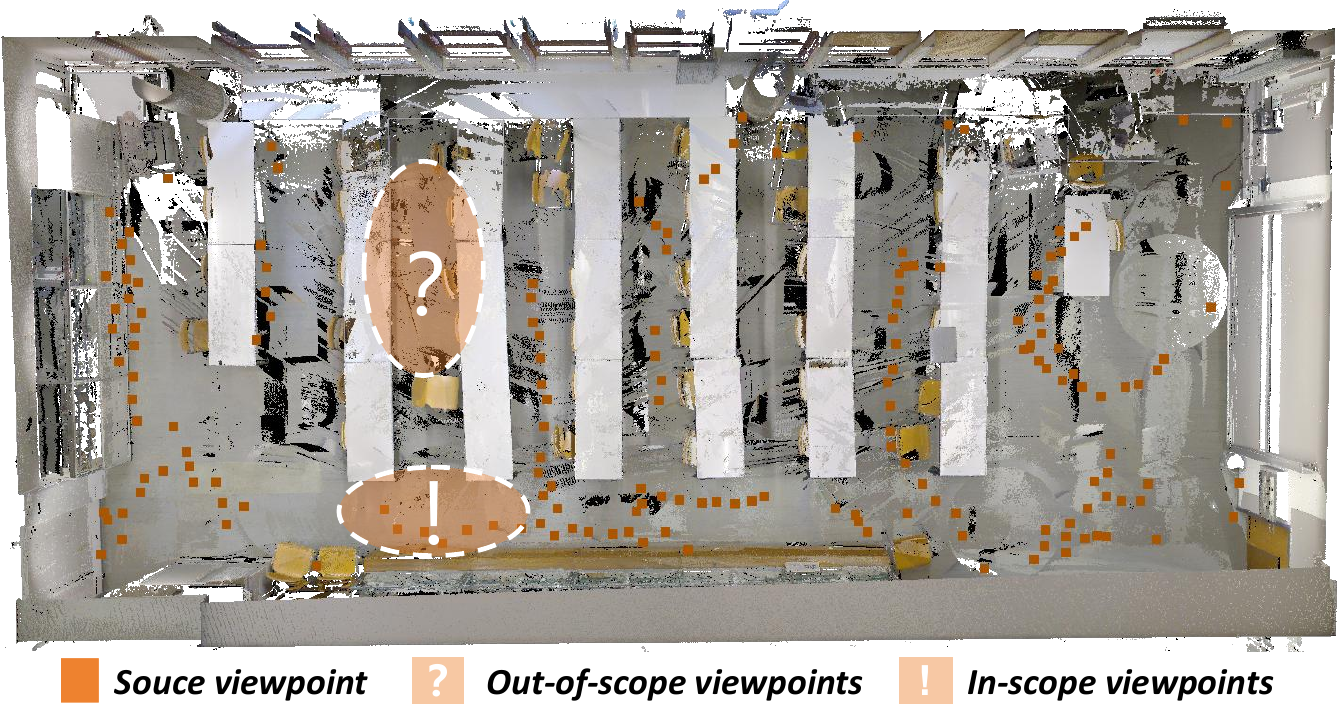}
	\end{minipage}
 
	\begin{minipage}[c]{0.23\textwidth}
		\includegraphics[width=\textwidth]{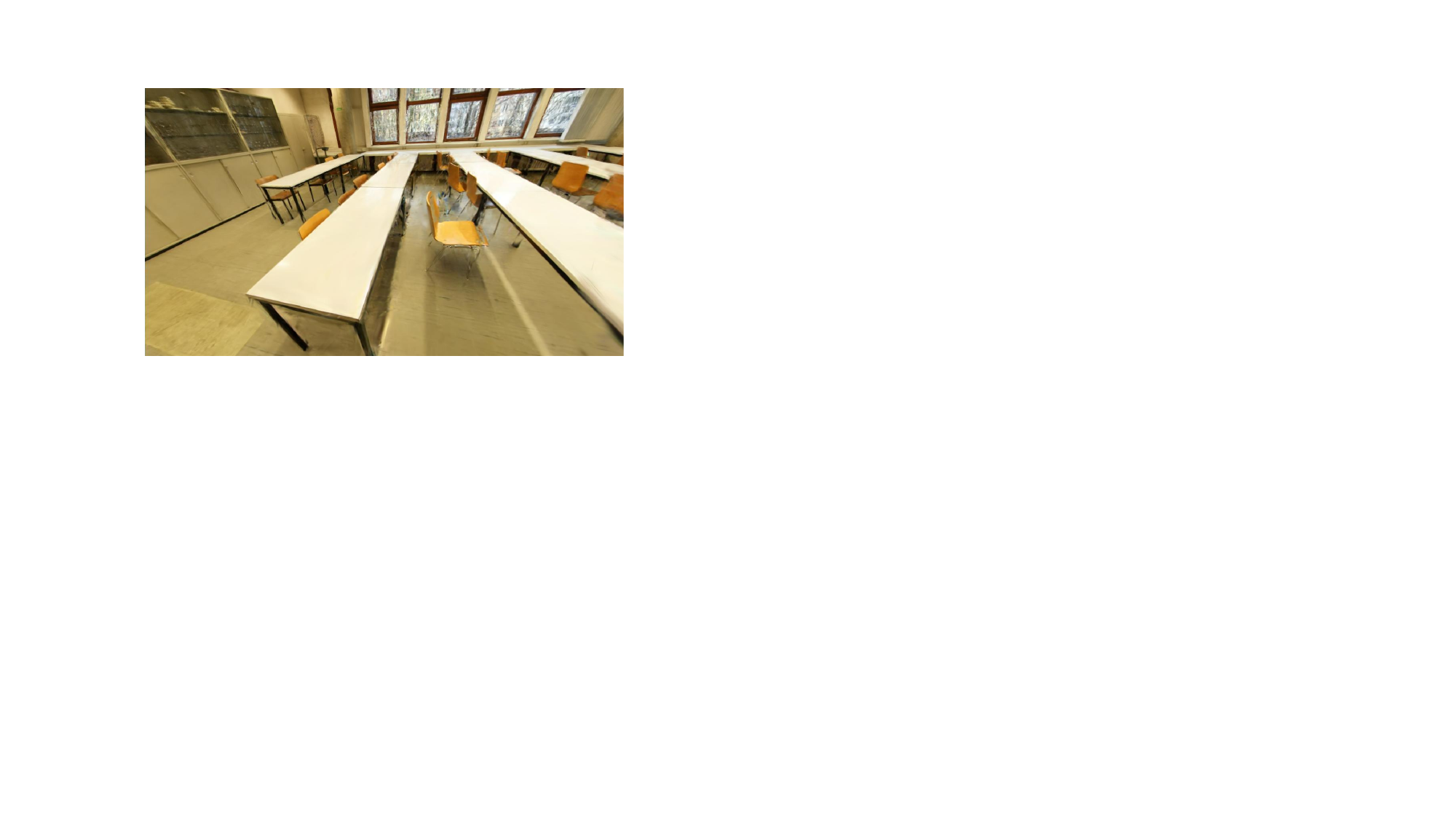}
		\subcaption{}
		\label{Baseline_1}
	\end{minipage} 
	\begin{minipage}[c]{0.23\textwidth}
		\includegraphics[width=\textwidth]{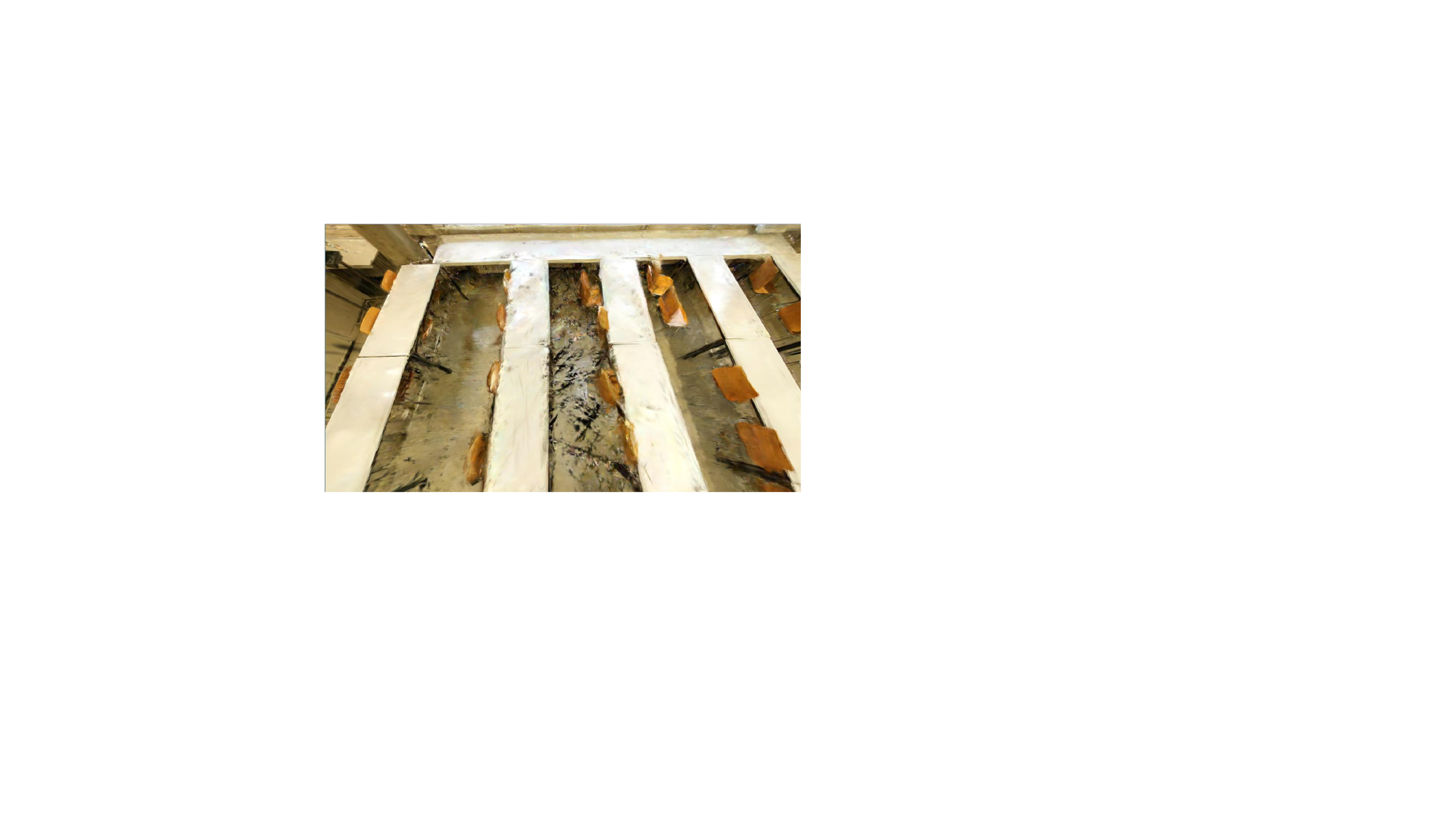}
		\subcaption{}
		\label{MIXED}
	\end{minipage}

	\caption{An instance of non-uniform observation. The top figure illustrates a scenario where the viewpoint in the question mark region spans across a wide baseline, while the exclamation mark region is within the narrow baseline. (a) and (b) show synthesized views for the same region when the viewpoint is positioned in the exclamation and question mark region, respectively. }
	\label{abstract}
\end{figure}

Unlike NeRF’s implicit 3D scene representation, 3D Gaussian Splatting (GS)~\cite{2} leverages Structure-from-Motion (SfM)~\cite{33} to generate a rough point cloud, representing the entire scene with Gaussian ellipses, which significantly reduces scene optimizing and rendering time. However, 3D GS depends heavily on various perspectives and is prone to overfitting in weakly observed regions, resulting in notable degradation in rendering quality~\cite{6, 34} and introducing potential reliability issues. Additionally, due to the movable range disparity between source image capture and scene visiting, even with increased shot density~\cite{31,8}, eliminating localized artifacts remains challenging.

\begin{figure*}[htp]
\centering
\captionsetup{type=figure}
\includegraphics[width=1.0\textwidth]{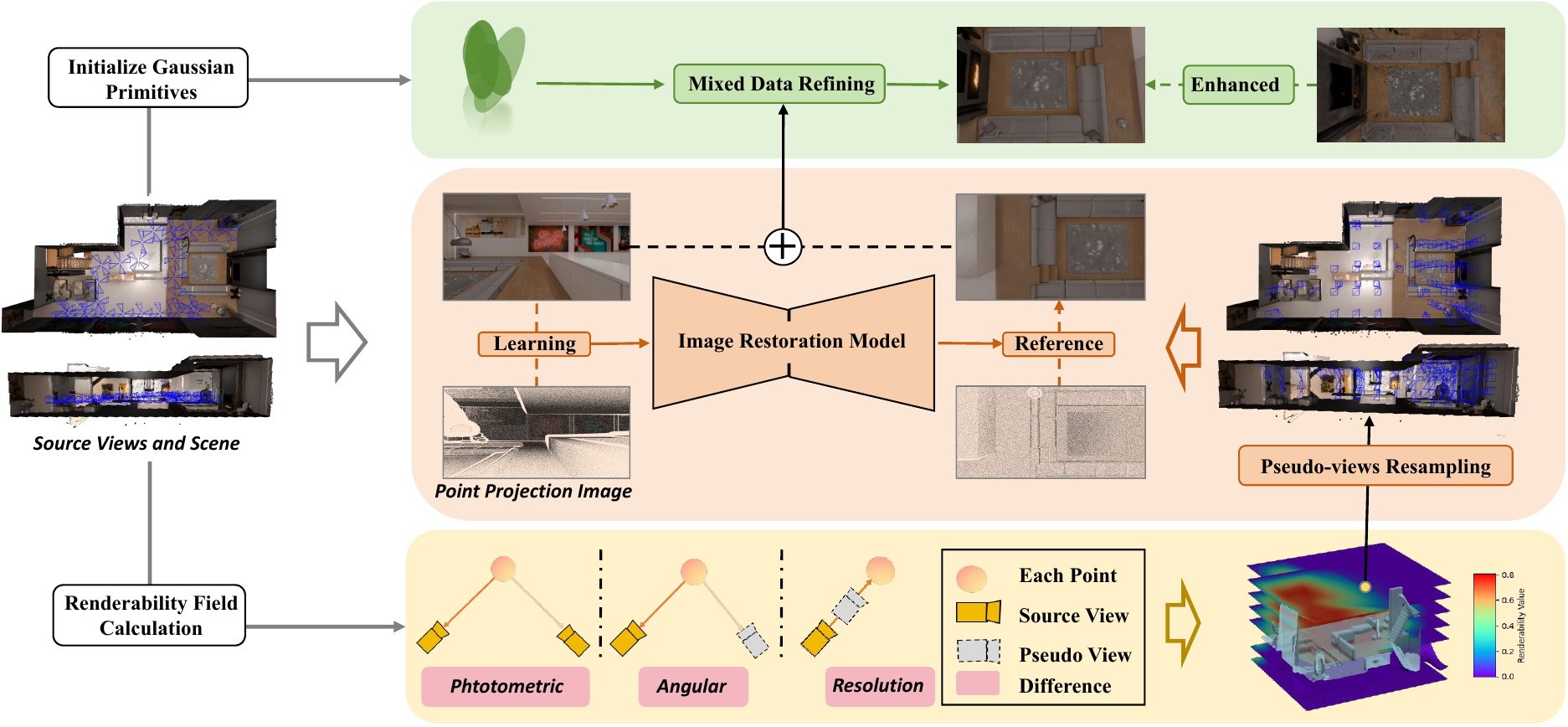}
\captionof{figure}{Overall pipeline of RF-GS. Given a series of source views and the scene point cloud, our method first calculates three metrics to obtain the renderability value of each viewpoint, which is then resampled to obtain the pseudo-view. The image restoration model is trained based on the source views, and the point cloud projection of the pseudo-view is restored to visible light image during inference. Finally, 3D Gaussian primitives are refined with the mixed data, significantly alleviating rendering artifacts compared to using only ground truth.}
\label{fig:structurel}
\end{figure*}

Non-uniform observations, characterized by low frequency of observations and incomplete directional coverage, pose significant challenges for stable view synthesis. Wide-baseline novel views differ substantially in viewing angle and position from the source views, often leading to weak observation. Fig.~\ref{abstract} indicates that, under wide baseline conditions, the rendering quality in non-uniform observation regions (e.g., the ground) is significantly lower than in uniform observation regions (e.g., the leftmost table) and diverges notably from the synthesized view produced under narrow baselines. To mitigate the negative effects of non-uniform inputs on the 3D GS model, we propose a renderability field-guided 3D GS approach, illustrated in Fig.~\ref{fig:structurel}. The core of our method is data augmentation through pseudo-views, which strengthens the model’s generalization capabilities.


Our contributions can be summarized as follows.

\textbullet~We propose a renderability field-guided pseudo-view sampling method that quantifies the distribution of scene observations and enhances perspective coverage in sparsely observed regions.

\textbullet~We address the challenge of wide baseline data augmentation by transforming geometrically consistent point-projection images into color images. This is achieved by training and using an image restoration model.

\textbullet~We discuss the limitations of current rendering quality evaluation methods for scene-oriented novel view synthesis and propose a new scheme to assess rendering stability.

\section{Related Works}
\label{sec:formatting}

High-fidelity real-time scene rendering has become crucial for various applications, including virtual reality (VR), autonomous driving, and digital twins. This demand has driven a surge in research to enhance the quality of rendering techniques to support these rapidly evolving fields.


\textbf{Scene representation.} Novel view synthesis focuses on generating novel views of a scene from a given set of images~\cite{35}. Mip-NeRF~\cite{9} improved the accuracy of high-frequency texture representations and the realism of generated images by refining the neural network structure, optimizing the loss function, and incorporating advanced and innovative training techniques. 

Other research efforts, including Tensorf~\cite{19}, Plenoxels~\cite{20}, and Instant-NGP~\cite{21}, have prioritized faster algorithm performance, often by reducing network complexity, employing more efficient optimization algorithms, or designing streamlined rendering pipelines. 3D GS~\cite{2} leveraged an explicit model to enable high-resolution real-time rendering. However, this effect relies on dense input data~\cite{3}: when the source views are sparse, 3D GS, like NeRF, faces challenges with generalizing well to unseen views synthesis~\cite{6}.


\textbf{View dependency.} Sparse inputs, while reducing data preparation costs, increase the risk of overfitting. RegNeRF~\cite{28} addressed this by estimating underlying scene geometry errors and applying careful regularization to enhance the appearance of unobserved views, followed by optimizing the ray-sampling space with an annealing strategy and refining the colors of unseen views using a normalized flow model. Diet-NeRF~\cite{4} sought to leverage a vision-language model~\cite{5} for rendering unseen views, though the high-level semantic guidance falls short in effectively aiding low-level reconstruction. Dreamfusion~\cite{9} introduced Score Distillation Sampling (SDS), utilizing 2D priors from pre-trained diffusion models to improve scene understanding and reconstruction. Methods like DiffusioNeRF~\cite{15}, SparseFusion~\cite{16}, and ReconFusion~\cite{17} also integrated diffusion models with NeRF, though they require extensive 3D data as priors, limiting their applicability.

FSGS~\cite{34} pioneered sparse view reconstruction for 3D Gaussian Splatting (GS) using a 2D depth network and augmented views to refine geometry. SparseGS~\cite{6} and DNGaussian~\cite{3} also relied on pre-trained 2D models for geometric accuracy, integrating SDS loss to enhance novel view synthesis. GaussianObject~\cite{23} employed view interpolation and a diffusion-based fine-tuning model~\cite{36} to improve pseudo-view rendering and multi-view consistency. Deceptive-NeRF/3DGS~\cite{49} addressed sparse-view reconstruction by sampling pseudo-viewpoints within the bounding box of initial viewpoints. They fundamentally limit the sample range to the original capture distribution rather than the entire scene. Feed-forward methods~\cite{12,44,45} adopt a two-stage approach—initialization and GS refinement—to enhance rendering under sparse inputs but require extensive pre-training on similar-scale scenes. 

\textbf{Uniform observations.} Novel view synthesis relies on angular uniformity and observation frequency~\cite{48}, which are often inconsistent in real-world data acquisition. Regularization-based methods struggle with non-uniform inputs, while interpolation-based augmentation fails under wide baselines, making stable rendering of arbitrary views in uncontrolled scenes challenging.

To enhance the generalization of 3D Gaussian Splatting (GS) models for applications such as VR, we introduce three key insights. First, observation non-uniformity can be quantified through the renderability field, and when combined with a pseudo-view sampling strategy, it enables precise viewpoint enhancement. Second, wide-baseline pseudo-view generation follows a denoising task based on 3D point projection images, which can ensure geometric consistency. Third, scene-level generalization evaluation can be validated through the construction of dense test cases.

\section{Methodology}

Given a scene containing a set of $N$ source views $I=\left\{I_{i}\right\}_{i=1}^{N}$ which is corresponding to real data, camera extrinsic $T=\left\{T_{i}\right\}_{i=1}^{N}$, intrinsics $C=\left\{C_{i}\right\}_{i=1}^{N}$, and a point cloud map $M$ obtained by the composition and merging of multiple LiDAR scans unified in its coordinate system, our goal is to sample pseudo-views $I_{p}$ through the renderability field to supplement the wide-baseline views. This process yields a 3D Gaussian model $M_{g}$, capable of generalizing to any view through optimization based on the hybrid data, thereby maintaining global rendering stability.

\subsection{Renderability Field Construction}\label{RFC}
The renderability field predicts the rendering quality for any viewpoint within the movable range of the reconstructed target and automatically identifies and prioritizes views that require enhancement.

\textbf{Intensive Sampling.} Benefiting from the point cloud map, we can specify viewpoint sampling intervals with a realistic scale. To ensure a comprehensive evaluation of complex indoor scenes, we perform voxel sampling within the bounding box of the environment. Given the bounding box \( B \) with maximum coordinates \( B_{\text{max}} \) and minimum coordinates \( B_{\text{min}} \), and a uniform sampling interval \( S \) along any axis, the set of viewpoints \( \boldsymbol{P_v} \) can be expressed as:

\begin{equation}\label{view sampling}
\begin{split}
\boldsymbol{P_v} = \big\{ (x, y, z) \mid\, & x = B_{\text{min},x} + iS, \\
                                  & y = B_{\text{min},y} + jS, \\
                                  & z = B_{\text{min},z} + k\frac{S}{2}, \\
                                  & i, j, k \in \mathbf{Z}, \, \text{and} \, (x, y, z) \in B \big\}.
\end{split}
\end{equation}

Since the source views are more centrally distributed on the Z-axis, we increased the sampling density on that axis. For each viewpoint, we generate six outward viewing directions to the positive and negative axes along $x$, $y$, and $z$, thus achieving a full 360-degree  observation, ensuring a comprehensive renderability assessment.

\textbf{Source Images Calculation.} To compute the renderability value, we need to determine the source view information associated with each point in the point cloud map. Given the real-world capture pose \( T = \{R, t\} \) and intrinsic parameters \( K \), we project each 3D point onto the image plane of all source views to identify which views observe it.  The projection equations are as follows:

\begin{equation}\label{source_img}
\left\{
\begin{array}{l}
p' = K (Rp + t),\\
P(p, T, K) = \left( \frac{p'_x}{p'_z}, \frac{p'_y}{p'_z} \right).
\end{array}
\right.
\end{equation}

Here, \( p \) is a point in 3D space, \( R \) is the world-to-camera rotation matrix, \( t \) is the translation vector, \( K \) is the intrinsic parameter matrix, \( p' \) are the projected pixel coordinates, and \( P(p, T, K) \) represents the image plane coordinates of point \( p \) given the pose \( T \) and intrinsic parameters \( K \). At the end of this step, for each 3D point, we obtain a set of source views that include it, along with the corresponding image indices, pixel positions, and color values. This information is crucial for estimating renderability in the pseudo-view image space.

\begin{figure}[!t]
\centering
\includegraphics[width=0.45\textwidth]{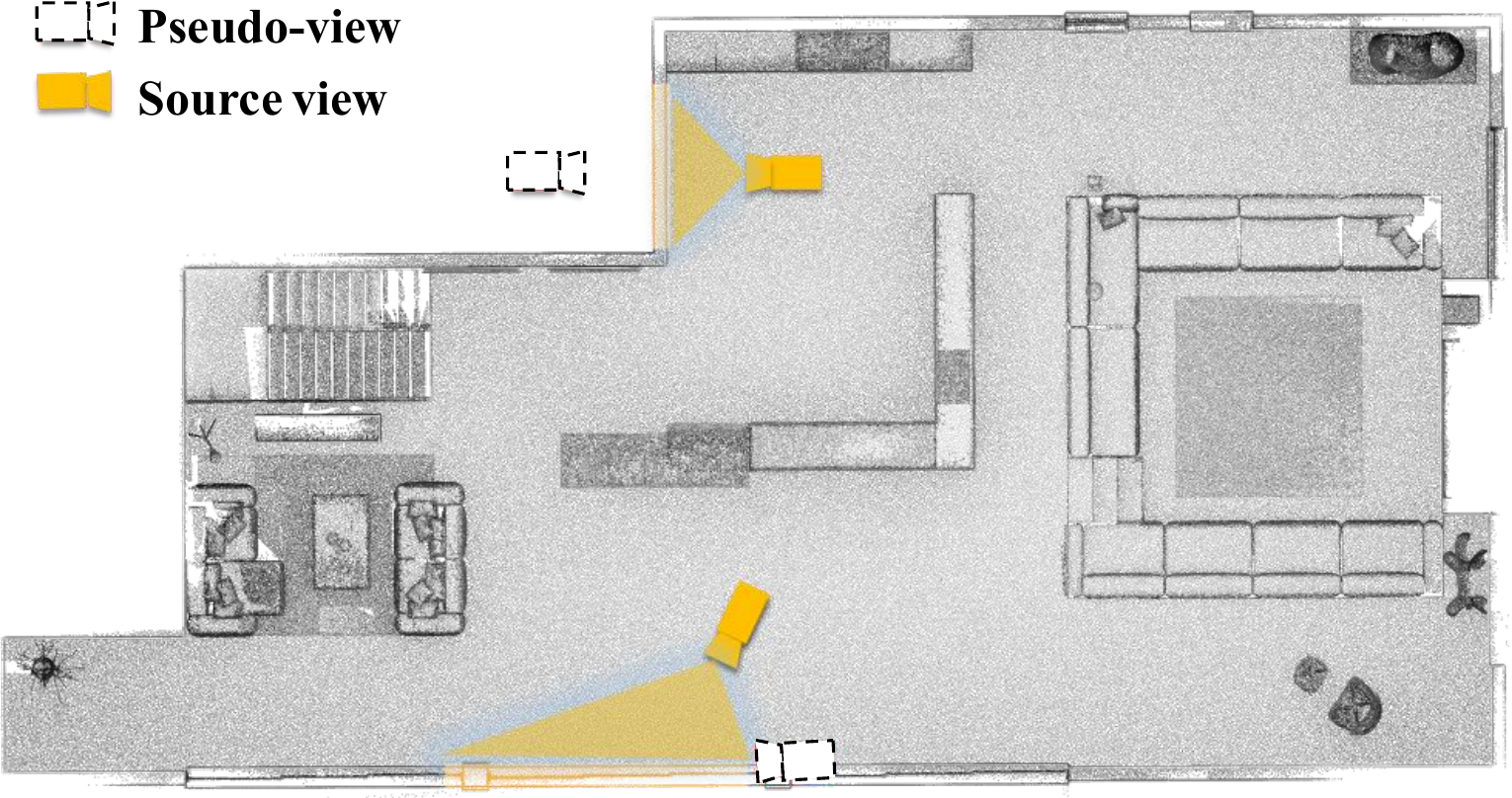}
\caption{Observation Conflict Issue. The point cloud captures only the scene’s internal structure, while pseudo-views observe external surfaces to establish co-visibility with source views. However, if no source view captures this exterior region, the co-visible area is falsely inferred.}
\label{fig_7}
\end{figure}
\textbf{Observation Conflict Judgement.} Since the pseudo-views are sampled from the scene's enclosing box, there may be meaningless views that are too close to the scene's point cloud or erroneous views that are not within the source view's viewing area. As shown in Fig.~\ref{fig_7}, since they do not establish a correct light reflection relationship with the source view, calculating the renderability value is unnecessary. 

To minimize the computation of irrelevant viewpoints, we implemented a filtering strategy to exclude them. If a pseudo-viewpoint has no unidirectional observation with any source viewpoints, it is in an unobserved area. We identify unidirectional observations through grid collision detection: for each pseudo-view, we first obtain the set of source views associated with it through the point cloud. 

Then, the scene point cloud undergoes hidden point removal based on the pseudo-viewpoints, resulting in a sub-map that is voxelized. By linearly interpolating a series of 3D points along the line connecting the pseudo-viewpoint and the source viewpoints, and checking whether these virtual positions (including the endpoints) lie inside the voxels, we determine if there is an occlusion between the viewpoints.

\textbf{Renderability Estimation.} We estimate the renderability based on the principles of image-based rendering (IBR). We introduce resolution, angular, and geometric metrics as proposed in previous work~\cite{41}. For any 3D point $p$ within a pseudo-view $I^i_p$, the geometric metric is represented by the average photometric difference between source views after normalizing the colors.
\begin{equation}\label{eq1}\left.H^p_{geo}=\left\{\begin{array}{ll}1-\frac{\sum_{v_i,v_j\in \boldsymbol{I_s},i\neq j}||c(v_i,p)-c(v_j,p)||_2}{\sqrt{3} \cdot \frac{N \cdot (N-1)}{2}},&N > 1,\\1,&N \leq 1.\end{array}\right.\right.\end{equation}

Here, $N$ denotes the number of source views $\boldsymbol{I_s}$ for point $p$, and $c(v_i,p)$ denotes the color of $p$ in $i$-th view. This metric shows the complexity of light change in the observation area. When considering variations in lighting, the angle metric \( H_{\text{ang}} \) additionally takes into account the minimum angle between the vectors formed by the co-visible point \( p \) with the pseudo-view \( I^i_p \) and any source view. Similarly, the resolution metric \( H_{\text{res}} \) additionally considers the minimum difference in distances from the co-visible point \( p \) to the pseudo-view \( I^i_p \) and any source view.

\begin{equation}\label{eq3}
\left\{
\begin{aligned}
H^p_{res} &= \exp\left(-\tan\left(\frac{\pi}{2} \cdot (1 - H^p_{geo})\right) \cdot \min_{o^i_s \in \boldsymbol{I_s}} d^s_p\right), \\
d^s_p &= \max \left(0, \frac{\|o^i_s - p\|_2 - \|o^i_p - p\|_2}{\|o^i_s - p\|_2}\right).
\end{aligned}
\right.
\end{equation}

\begin{equation}\label{eq2}\begin{aligned}H^p_{ang}=\exp\left(-\tan{(\frac{\pi}{2}\cdot (1-H^p_{geo}))}\cdot\min_{o^i_s\in \boldsymbol{I_s}}\angle{o^i_ppo^i_s}\right).\end{aligned}\end{equation}

Let \( M^i_p \) represent the point cloud under the pseudo-view \( I^i_p \), and let \( o_s \) and \( o_p \) denote the coordinates of the source viewpoint and the pseudo-view point, respectively. $d^s_p$ denotes the distance difference between the pseudo-viewpoints and the source viewpoints to the co-visible point, which describes the resolution disparity. The comprehensive metric $V^i_p$ for the final view $I^i_p$ can be expressed as the product of the average values of each term:

\begin{equation}\label{eq4}\begin{aligned} V^i_p = \overline{H^i_{\text{geo}}} \cdot \overline{H^i_{\text{res}}} \cdot \overline{H^i_{\text{ang}}}.\end{aligned}\end{equation}

Since the renderability value directly reflects the rendering quality, we can filter the pseudo-views by preset parameters.

\subsection{Point Projection Image Restoration}

Image restoration models often rely on contextual inference to fill missing regions but lack 3D spatial awareness, leading to geometric inconsistencies. Additionally, noise in the initial image can adversely affect restoration outcomes. Incorporating low-quality pseudo-views during training may further degrade the reconstructed scene.

Point projection images, derived from discrete 3D points, inherently ensure geometric consistency and multi-view generalizability, thereby reducing the complexity of obtaining wide-baseline view information. Consequently, we utilize point projection images as the initial input. We train the Nonlinear Activation Free Network (NAFNet)~\cite{21}, a deep learning model designed for image restoration, to infer RGB images from these point projection images. NAFNet is straightforward to train and deploy, demonstrating high effectiveness for this task.

\begin{figure}[t]
\centering
\includegraphics[width=0.45\textwidth]{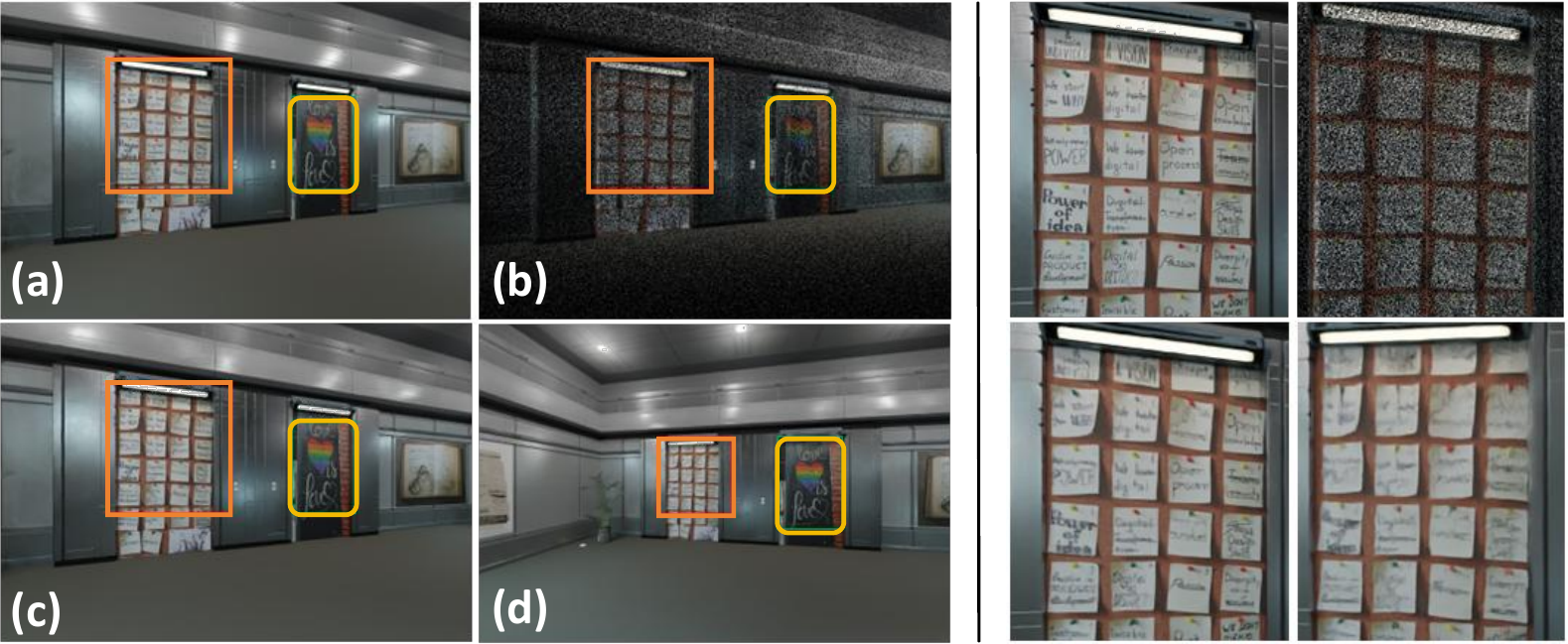}
\caption{Point cloud image restoration. (a) is the ground truth, (b) is the point cloud image from the viewpoint aligned with the ground truth, (c) the restored image corresponding to the point cloud image in the ground truth view, and (d) the restored image from a new wide baseline viewpoint. The four images on the right display zoomed-in regions, highlighted within the orange boxes in images (a) through (d).}
\label{fig_8}
\end{figure}

During training, since inputs consist of discrete points (as shown in Fig.~\ref{fig_8}b), we treat these images as noise-laden, using the actual captured image as a reference to learn the mapping from the noisy point projection image to the real image, following a denoising task.

In the inference stage, we use the previously obtained pseudo-view pose to render the point projection image, then apply the model to infer the corresponding color image, as shown in Fig.~\ref{fig_8}d. This output consistently preserves scene geometry and weakly textured regions.

\subsection{Staged Gaussian Primitive Optimization}\label{batch}
Considering that high-frequency texture regions cannot be recovered in pseudo-views (Fig.~\ref{fig_8}), we propose a refining strategy in two stages to mitigate the effects of pseudo-views.

In the first stage, we mix all the data and treat the pseudo-view as a noisy ground truth (GT) view, using its angular information to approximate the properties of Gaussian primitives with a high learning rate to better match the actual observations. In this process, the pseudo-view is restricted from guiding the densification operation, as blurred textures could lead to an unintended merging of geometric primitives. 

In the second stage, we fine-tune using only the real images to recover geometric edges and textures by optimizing the spherical harmonic and freezing other elements. As a result, the overall training loss function is similar to vanilla GS, which is defined as follows:
\begin{equation}\label{eq5}
 \mathcal{L} = \lambda_{rgb} \mathcal{L}_1(\hat{I}_i, I_i) + \lambda_{ssim} \mathcal{L}_{ssim}(\hat{I}_i, I_i),
\end{equation}

where $\mathcal{L}_1$ is the photometric loss and $\mathcal{L}_{ssim}$ is the structural similarity loss.


\begin{table*}[!t]
\centering
\caption{Quantitative results on synthetic data. Thirteen views were separated from captures for narrow baseline assessment, while 5,449 views were intensively sampled from the scenes. Optimal results are in bold, with secondary results underlined.}
\begin{tabular*}{\textwidth}{@{\extracolsep{\fill}} c|cccc|cccc }
\hline
\multicolumn{1}{l|}{} & \multicolumn{4}{c|}{13 views}                                       & \multicolumn{4}{c}{5,449 views}                                         \\
Method                & PSNR $\uparrow$        & SSIM $\uparrow$         & LPIPS $\downarrow$         & SDP $\downarrow$         & PSNR $\uparrow$         & SSIM $\uparrow$         & LPIPS $\downarrow$         & SDP $\downarrow$         \\ \hline
Vanilla GS~\cite{2}            & \textbf{34.3} & \ul{0.963}          & \ul{0.116}          & \ul{2.26}          & \ul{29.67}          & \ul{0.927}          & \ul{0.174}          & 4.97          \\
CoR-GS~\cite{42}                & \ul{34.29}         & \textbf{0.964} & \textbf{0.113} & \textbf{2.04} & 28.68          & 0.921          & \textbf{0.171}          & 6.12          \\
DNGaussian~\cite{3}                & 30.22         & 0.943          & 0.177          & 2.36          & 27.4           & 0.908          & 0.22           & 4.62          \\
FSGS~\cite{34}                  & 29.82         & 0.945          & 0.176          & 2.69          & 26.94          & 0.913          & 0.215          & 4.62          \\
SparseGS~\cite{6}             & 28.67         & 0.94           & 0.167          & 3.96          & 27.39          & 0.916          & 0.205          & \ul{4.36}          \\
Octree-GS~\cite{47}             & 33.13         & 0.956          & 0.131          & 2.43          & 28.68          & 0.915          & 0.201          & \textbf{4.27} \\
RF-GS (Ours)           & 33.62         & 0.962          & 0.123          & 2.3           & \textbf{29.97} & \textbf{0.933} & \textbf{0.171} & 4.61          \\ \hline
\end{tabular*}
\label{limitation}
\end{table*}
\section{Experiments}
\subsection{Methods of Comparison}
\indent We compare our method with five approaches, where CoR-GS~\cite{42} regularizes inconsistencies across multiple renderings, while DNGaussian~\cite{3} enforces geometric consistency using depth and normal. SparseGS~\cite{6} leverages pre-trained diffusion models for pseudo-view supervision. FSGS~\cite{34} guides Gaussian densification to improve unseen view rendering, and Octree-GS~\cite{47} ensures high-quality synthesis with well-initialized  Gaussian primitives.


\subsection{Dataset Preparation}

We prepared a diverse set of indoor and outdoor datasets for quality comparisons of novel view synthesis in larger-scale scenes. The synthetic data was generated from building models in Blender, using a multi-camera setup to capture 360° views per viewpoint. Point cloud data was obtained by sampling discrete 3D points on the mesh model. The capture path simulates real data acquisition, validating the training strategy's effectiveness and assessing limitations in prior metrics. The real-world data is categorized into indoor and outdoor scenes. The indoor data originates from ScanNet++ \cite{46}, from which we selected a few of the largest scenes. Outdoor data was collected using a customized mobile LiDAR system with two PandarXT32 LiDAR sensors and four GoPro Hero10 cameras. These data serve to evaluate the effectiveness of the algorithm.

For test data, we applied the farthest point sampling (FPS) method to the camera viewpoints in the synthetic dataset, following a standard test-to-training ratio of 1:7 to ensure uniform spatial distribution of test data. Additionally, we used the sampling method from Section~\ref{RFC} on the scene with a step size $S=1 m$, resulting in 5,449 images for testing the ground truth of the algorithm's performance.

\subsection{Metrics of Evaluation}
To evaluate our approach, the average peak signal-to-noise ratio (PSNR), structural similarity index measure (SSIM)~\cite{28}, and learned perceptual image patch similarity (LPIPS)~\cite{35} were measured. In addition, we express the generalization of the novel perspective in terms of the standard deviation of the PSNR (SDP) of the test data.

\subsection{Details of Experiments}
The method was tested on a single NVIDIA RTX 4080S GPU with CUDA 11.8. Pseudo-viewpoint sampling was set to $S=2$ m for simulated data, $S=6$ m for real-world data, and $S=1$ m for the smaller public dataset. The pseudo-view resampling ranges are all set to $[0.1,0.6]$. NAFNet training parameters follow that in~\cite {21}, and GS model training parameters match~\cite{2}, with 20,000 and 10,000 iterations for hybrid and real images-only data, respectively. All methods were initialized using the LiDAR-generated map to ensure fair, consistent inputs.

\begin{figure}[t]
        \centering
	\begin{minipage}[c]{0.47\textwidth}
		\includegraphics[width=\textwidth]{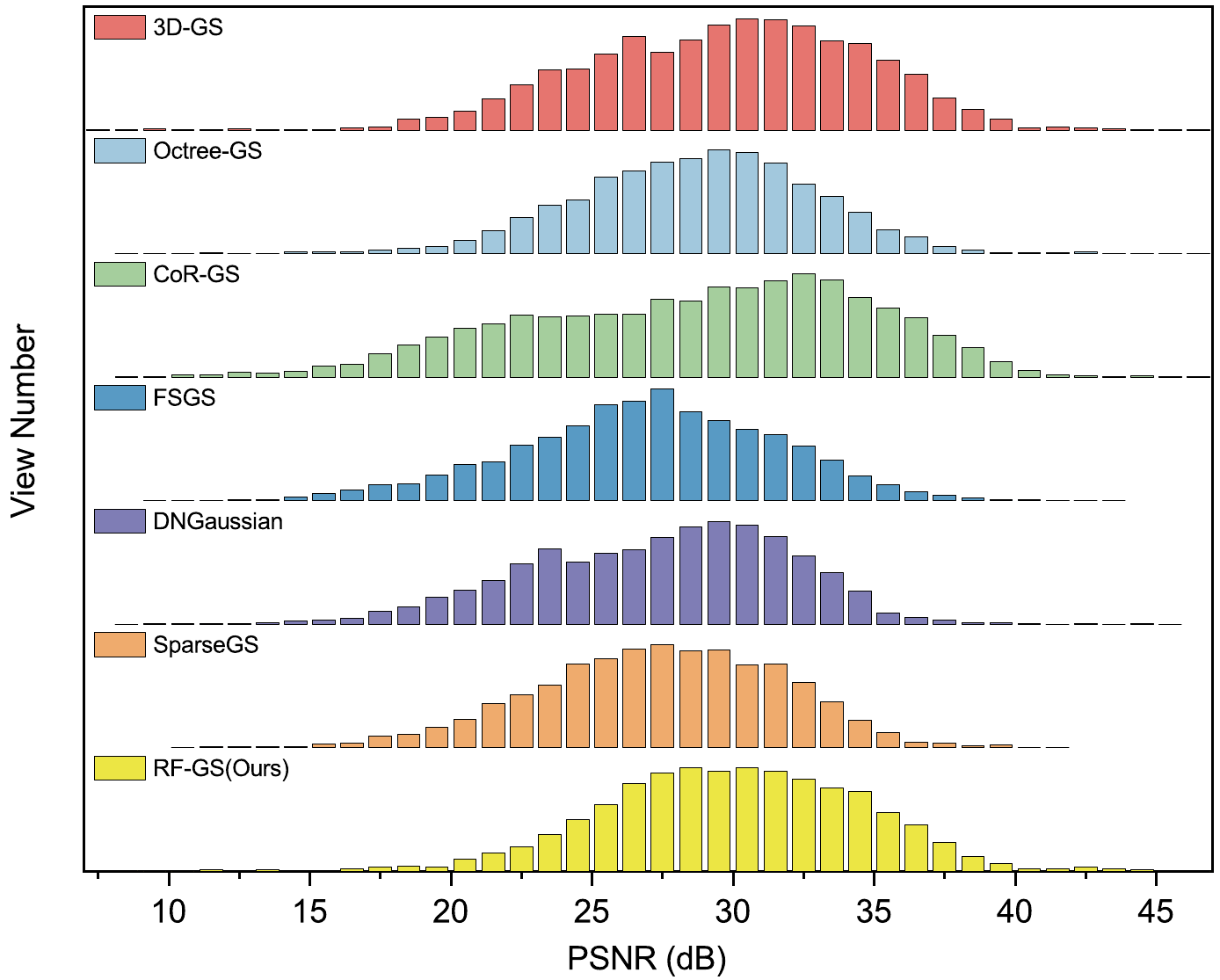}
	\end{minipage}
	\caption{Distribution of PSNR in simulated data. The figure shows the distribution of PSNR across 5449 test cases in a dense simulation scenario, with each algorithm marked by a distinct color. The x-axis represents PSNR values with a scale of 1, and the y-axis shows the number of test cases falling within each bin. }
	\label{PSNR_DIS}
\end{figure}

\begin{figure*}[htp]
        \centering
	\begin{minipage}[c]{0.24\textwidth}
		\includegraphics[width=\textwidth]{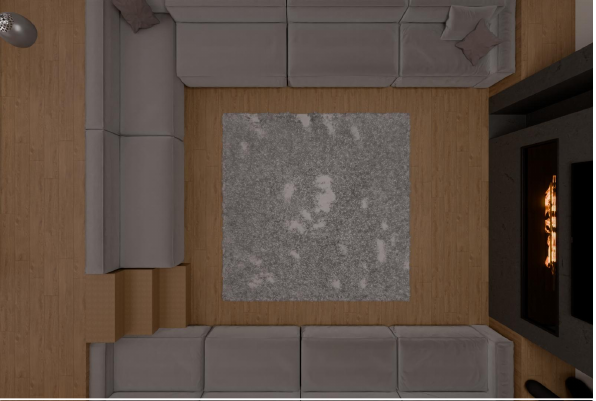}
		\subcaption{GT}
		\label{GT_2}
	\end{minipage}
	\begin{minipage}[c]{0.24\textwidth}
		\includegraphics[width=\textwidth]{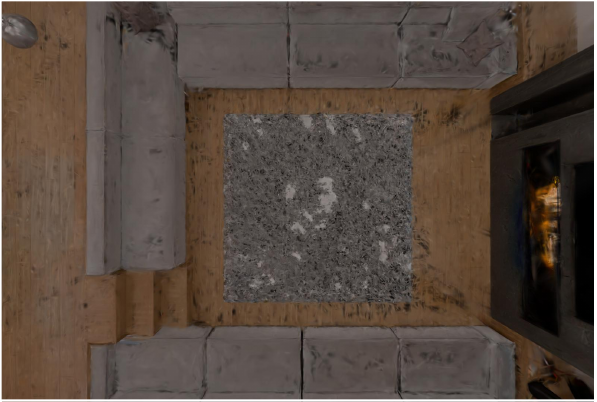}
		\subcaption{Vanilla GS \cite{2}}
		\label{Baseline_2}
	\end{minipage} 
	\begin{minipage}[c]{0.24\textwidth}
		\includegraphics[width=\textwidth]{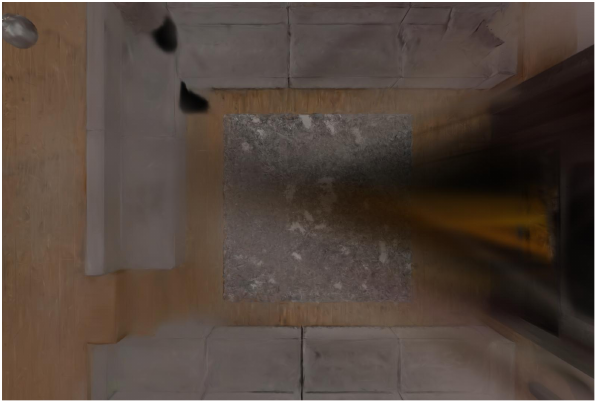}
		\subcaption{CoR-GS \cite{42}}
		\label{CoR}
	\end{minipage}
 	\begin{minipage}[c]{0.24\textwidth}
		\includegraphics[width=\textwidth]{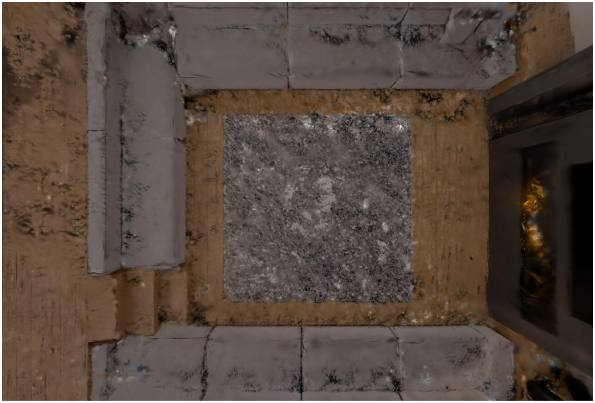}
		\subcaption{DNGaussian \cite{3}}
		\label{DNGgaussian}
	\end{minipage}
 
  	\begin{minipage}[c]{0.24\textwidth}
		\includegraphics[width=\textwidth]{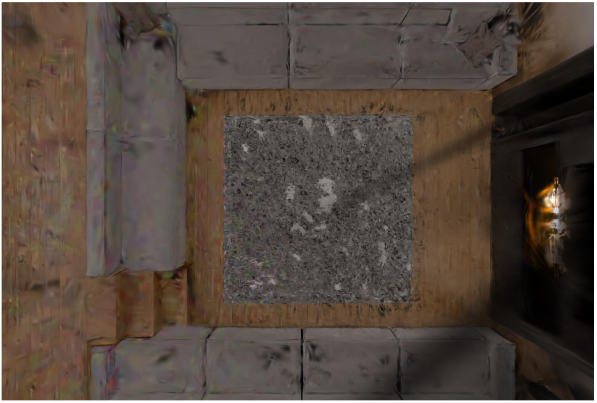}
		\subcaption{SparseGS \cite{6}}
		\label{Sparse-GS}
	\end{minipage}
   	\begin{minipage}[c]{0.24\textwidth}
		\includegraphics[width=\textwidth]{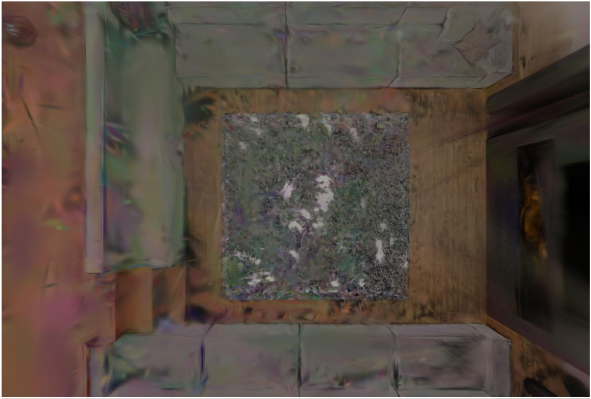}
		\subcaption{FSGS \cite{34}}
		\label{FSGS}
	\end{minipage}
        \begin{minipage}[c]{0.24\textwidth}
		\includegraphics[width=\textwidth]{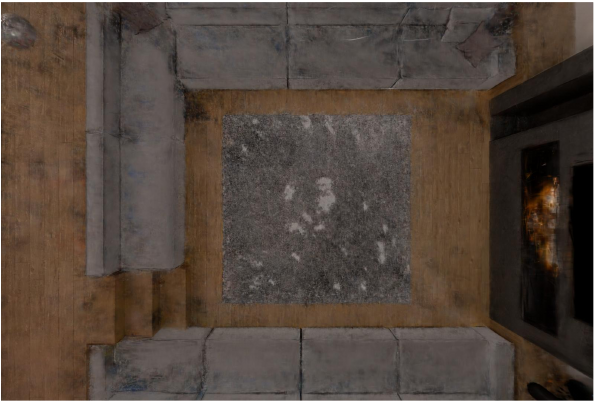}
		\subcaption{Octree-GS \cite{47}}
		\label{Octree-GS}
	\end{minipage}
        \begin{minipage}[c]{0.24\textwidth}
		\includegraphics[width=\textwidth]{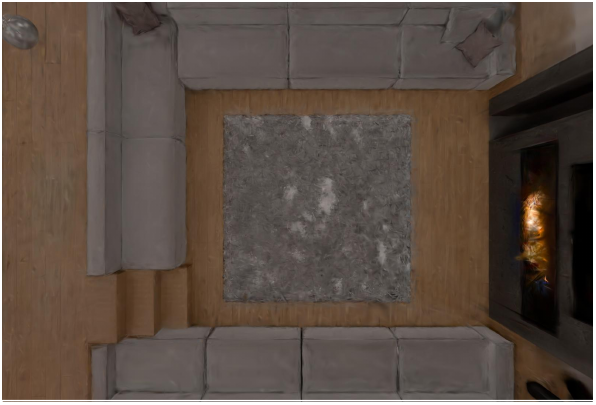}
		\subcaption{RF-GS (Ours)}
		\label{RF-GS}
	\end{minipage}

	\caption{View synthesis on the synthetic data. The view is a top view, which differs significantly from the training view. As revealed in comparisons of our model to other GS methods, RF-GS produces more realistic results. Our method mitigates the problems of incomplete, blurring, and numerous artifacts present in new-view synthesis across wide baselines.}
	\label{limitation fig}
\end{figure*}

\section{Results}

\subsection{Synthetic Data}

\textbf{Limitations of evaluation method.}  As shown in Table~\ref{limitation}, we take the evaluation results from 5,449 views as the ground truth for algorithm performance, while the results from 13 views represent the performance prediction. Although FSGS outperforms SparseGS in local evaluations, this conclusion does not match the global evaluations, indicating that local evaluation results are unreliable for representing the rendering quality of the whole scene. 

Additionally, the global evaluation indicates that CoR-GS matches Octree-GS in PSNR and outperforms it in SSIM and LPIPS. However, as shown in Fig.~\ref{CoR}, CoR-GS produces intolerable artifacts and is noticeably weaker than Octree-GS. Combined with the SDP analysis, CoR-GS demonstrates better rendering quality and stability within the local range. Still, its generalization declines significantly beyond this range, indicating more severe overfitting. 

Traditional metrics often represent average test values and may not capture quality fluctuations. To address this, we introduce SDP as a metric to assess the stability of algorithms when their performance across traditional metrics is similar. By combining all four metrics using spatially uniform test cases, we ensure a reliable evaluation of synthesis quality.

\begin{figure}[t]
        \centering
	\begin{minipage}[c]{0.23\textwidth}
		\includegraphics[width=\textwidth]{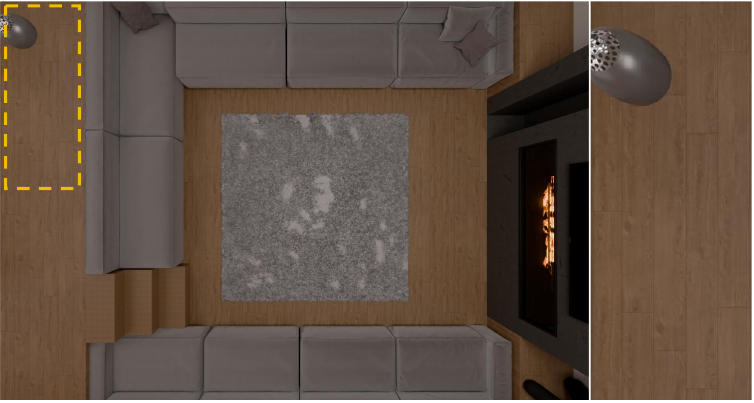}
		\subcaption{GT}
		\label{GT_3}
	\end{minipage}
	\begin{minipage}[c]{0.23\textwidth}
		\includegraphics[width=\textwidth]{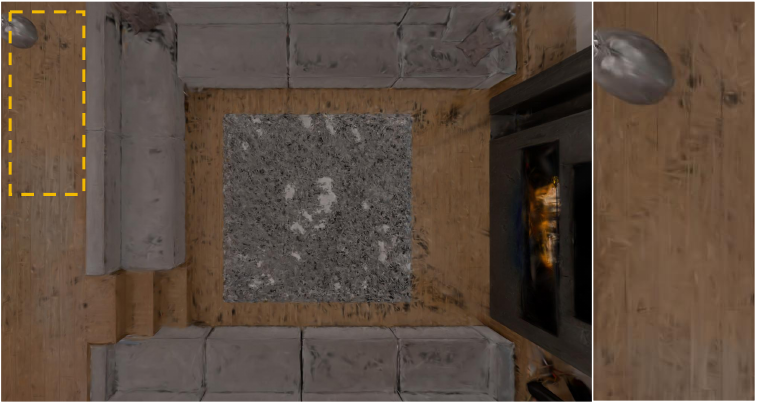}
		\subcaption{w/o $P$}
		\label{Baseline_3}
	\end{minipage} 
	\begin{minipage}[c]{0.23\textwidth}
		\includegraphics[width=\textwidth]{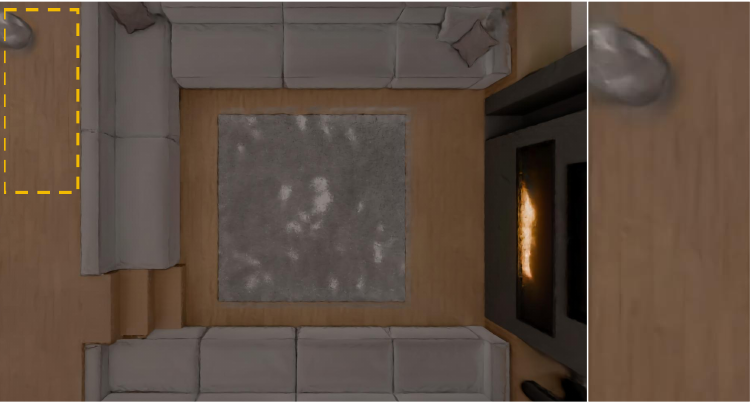}
		\subcaption{w/ $P$}
		\label{Mixed data}
	\end{minipage}
 	\begin{minipage}[c]{0.23\textwidth}
		\includegraphics[width=\textwidth]{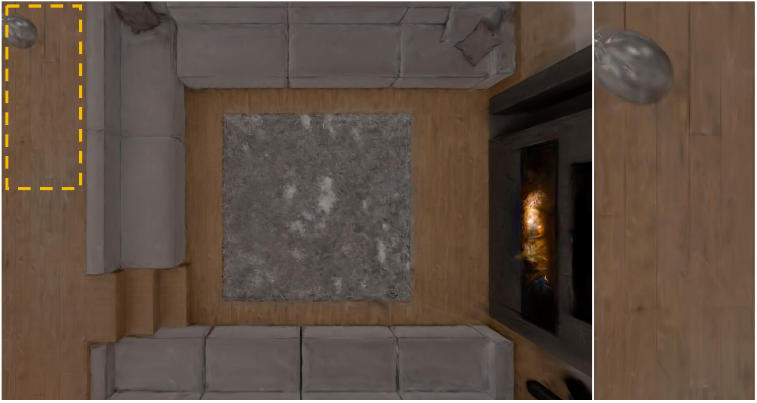}
		\subcaption{(w/ $P$, w/o $P$)}
		\label{2STAGE}
	\end{minipage}

	\caption{Effectiveness of staged optimization. The zoomed-in area highlights changes in rendering detail.}
	\label{batch train}
\end{figure}

\textbf{Advantages of the proposed method.} To support a free-roaming experience, the model must generalize across all possible viewpoints within the scene. Top views, which are out of scope, differ significantly from source views, amplifying Gaussian primitive misalignment and resulting in artifacts and voids (see Fig.~\ref{DNGgaussian}).
As shown in Fig.~\ref{PSNR_DIS}, using PSNR~$=~25$ as a reference line, our method reduces the number of low-quality views and raises the lower bound of novel view synthesis quality, while maintaining the high-quality view distribution of 3D-GS. The new perspectives rendered by Octree-GS are sparse in the range of PSNR~$>~35$, indicating a reduction in the upper bound of rendering quality. Other methods show a clear gap compared to ours.

\begin{table}[t]
\centering
\caption{Hybrid data training method. $P$ represents pseudo-views, where (w/ $P$, w/o $P$) indicates using mixed data in stage one, and only real captured data in stage two. $P^*$ denotes a pseudo-view obtained using random sampling not based on renderability values.}
\begin{tabular}{ccccc}
\hline
Method       & PSNR $\uparrow$        & SSIM $\uparrow$       & LPIPS $\downarrow$        & SDP $\downarrow$        \\ \hline
w/o $P$        & 29.67          & 0.927          & 0.174          & 4.97          \\
w/ $P$         & 28.75          & 0.924          & 0.191          & 4.52 \\
(w/ $P$, w/o $P$) & \textbf{29.97} & \textbf{0.933} & \textbf{0.171} & 4.61          \\ 
(w/ $P^*$, w/o $P^*$) & 29.31 & 0.926 & 0.190 & \textbf{4.45}          \\ \hline
\end{tabular}
\label{ablation}
\end{table}

\begin{figure*}[t]
        \centering
        \begin{minipage}[c]{0.49\textwidth}
	\begin{minipage}[c]{0.96\textwidth}
		\includegraphics[width=\textwidth]{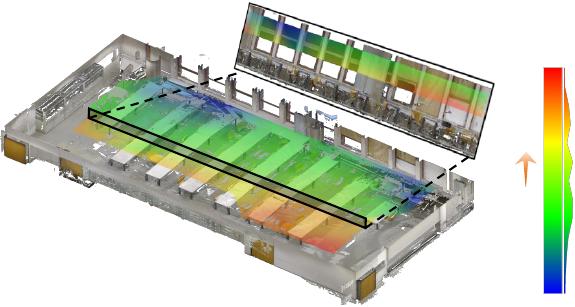}
	\end{minipage}
        \vspace{0.5cm}
        
	\begin{minipage}[c]{0.49\textwidth}
		\includegraphics[width=\textwidth]{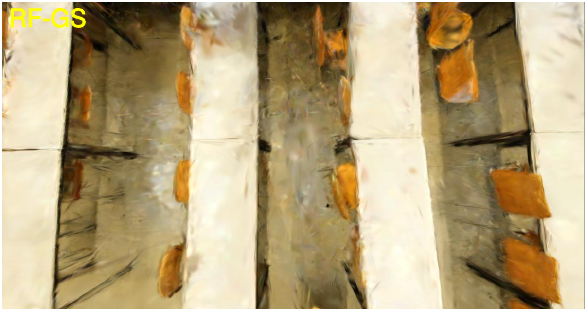}
	\end{minipage} 
	\begin{minipage}[c]{0.49\textwidth}
		\includegraphics[width=\textwidth]{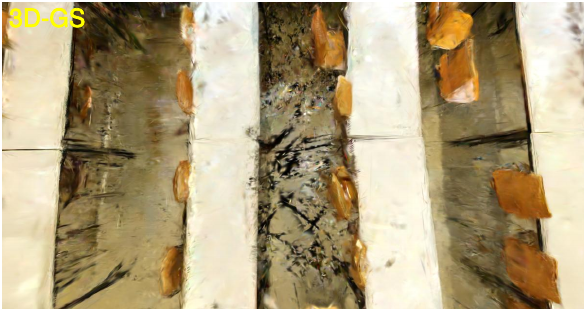}
	\end{minipage}
 
 	\begin{minipage}[c]{0.49\textwidth}
		\includegraphics[width=\textwidth]{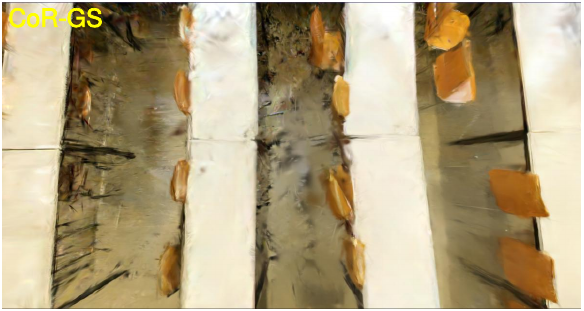}
	\end{minipage}
  	\begin{minipage}[c]{0.49\textwidth}
		\includegraphics[width=\textwidth]{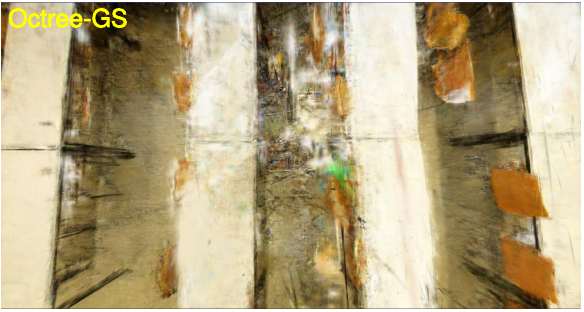}
	\end{minipage}
 		\subcaption{ScanNet++}
		\label{scannet}
        \end{minipage}
                \begin{minipage}[c]{0.49\textwidth}
	\begin{minipage}[c]{0.96\textwidth}
		\includegraphics[width=\textwidth]{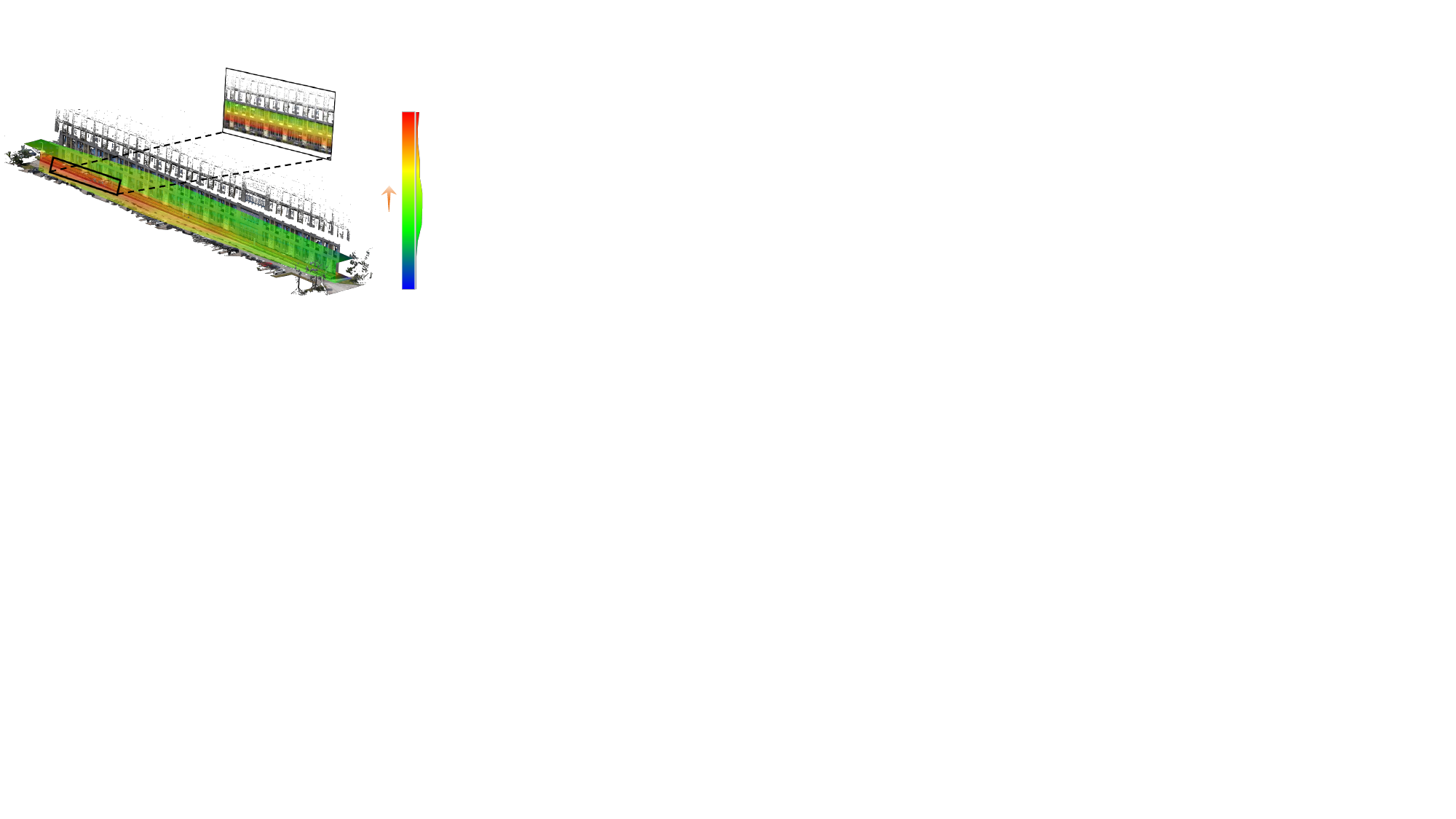}
	\end{minipage}
         \vspace{0.46cm}
 
	\begin{minipage}[c]{0.49\textwidth}
		\includegraphics[width=\textwidth]{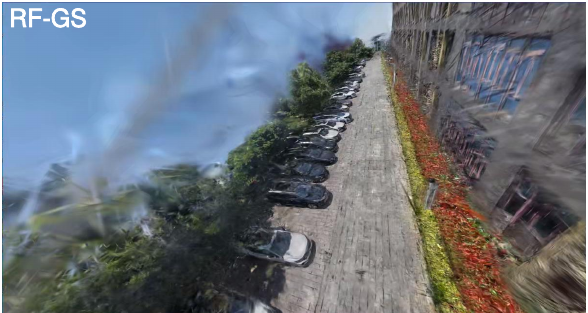}
	\end{minipage} 
	\begin{minipage}[c]{0.49\textwidth}
		\includegraphics[width=\textwidth]{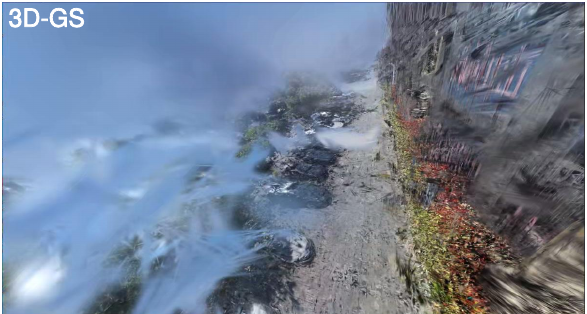}
	\end{minipage}

 	\begin{minipage}[c]{0.49\textwidth}
		\includegraphics[width=\textwidth]{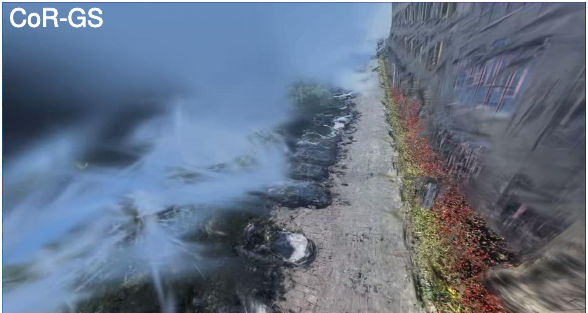}
	\end{minipage}
  	\begin{minipage}[c]{0.49\textwidth}
		\includegraphics[width=\textwidth]{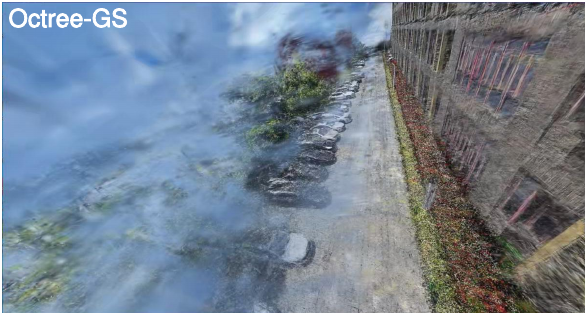}
	\end{minipage}
  		\subcaption{Custom}
		\label{realworld}
        \end{minipage}

	\caption{View synthesis on real-world data. The top image illustrates the scene along with its renderability field, where color transitions from blue to red indicate increasing renderability values. (a) and (b) show the results of wide-baseline synthesis for various algorithms applied to the corresponding scenes, where no ground truth is provided.}
	\label{real-world}
\end{figure*}

\begin{table*}[t]
\centering
\caption{Quantitative results on real-world data. The test data are sampled at a 1:10 ratio from the source views, yielding 17 samples from the ScanNet++ dataset and 16 from the custom dataset. The optimal results are marked in bold, and the secondary results are underlined.}
\begin{tabular*}{\textwidth}{@{\extracolsep{\fill}} c|cccc|cccc }
\hline
\multicolumn{1}{l|}{}          & \multicolumn{4}{c|}{ScanNet++}                                                                                                           & \multicolumn{4}{c}{Custom}                                                                                                           \\
Method                         & PSNR $\uparrow$                               & SSIM $\uparrow$                      & LPIPS $\downarrow$                              & SDP $\downarrow$                                & PSNR $\uparrow$                      & SSIM $\uparrow$                            & LPIPS $\downarrow$                              & SDP $\downarrow$                               \\ \hline
CoR-GS~\cite{42}                         & 25.93                              & \textbf{0.876}            & 0.192                              & 2.7                          &  17.28               & {\ul{0.485}}                           & {\ul{0.471}}                        & 1.73                              \\
DNGaussian~\cite{3}                         & 24.56          & 0.85  & 0.257          & {\textbf{2.17}} & 16.96 & 0.463       & 0.505          & {\ul{1.03}} \\
FSGS~\cite{34}                           & 24.71          & 0.856 & 0.245          & {\ul{2.65}}          & 16.28 & 0.452       & 0.581          & 1.05   \\
SparseGS~\cite{6}   & \textemdash{}                   & \textemdash{}          & \textemdash{}                   & \textemdash{}                   & \textemdash{}          & \textemdash{}                & \textemdash{}                   & \textemdash{}                  \\
Octree-GS~\cite{47}  & {\textbf{26.24}} & 0.87  & {\textbf{0.171}} & 3.08          & {\ul{17.6}}  &  0.469 & {\textbf{0.423}} & \textbf{0.729}         \\
Vanilla GS~\cite{2}                    & 25.93                              & 0.873                     & {\ul{0.191}}                        & 2.86                               & 16.65                     & 0.456                           & 0.491                              & 1.21                              \\
RF-GS (Ours)                          & {\ul{25.98}}                        & {\ul{0.874}}               & 0.215                              & 2.74                               & \textbf{18.49}            & \textbf{0.504}                  & 0.474                              & 1.09                              \\ \hline
\end{tabular*}
\label{real_table}
\end{table*}
\textbf{Effectiveness of staged Gaussian primitives optimization.}
Table~\ref{ablation} demonstrates the effectiveness of our strategy in Section~\ref{batch}.  Hybrid data training enhances generalization, achieving the lowest SDP but sacrificing some rendering quality. Fig.~\ref{Mixed data} shows that while it eliminates artifacts and ensures consistency, it results in overly smooth appearances, lacking realism in reflections and textures. To mitigate this, we fine-tune colors with ground truth data, refining details (Fig.~\ref{2STAGE}). This approach surpasses the baseline in both quality and generalization.

\subsection{Real-world Data}
Intensive testing is difficult to obtain in the real world compared to simulated data, so it must be supplemented with qualitative results to evaluate the algorithm's performance. 

ScanNet++ data is relatively dense, so the test data mainly quantifies in-scope view quality and has limited relevance to overall scene generalization. Table~\ref{real_table} indicates that RF-GS performs comparably to standard GS, suggesting that the added pseudo-views did not interfere with source view fitting. Combining with qualitative results, Fig.~\ref{scannet} demonstrates that our method significantly enhances challenging regions without introducing noticeable artifacts or voids, outperforming other methods.

The sparse custom data reflects method performance on wide baselines and partially represents generalization across the scene. Table~\ref{real_table} shows that our method achieves minimal distortion, strong geometric consistency, and good stability. As seen in Fig.~\ref{realworld}, RF-GS enables stable rendering from any viewpoint, whereas other methods exhibit intolerable blurring in challenging areas.

\section{Conclusion}

In this paper, we propose a method using renderability fields to enhance generalization in free-scene rendering. First, we introduce a renderability field to guide wide-baseline pseudo-view selection, intensifying supervision. Second, we generate color images from point-projected images via an image restoration model, ensuring both geometric consistency and image quality in pseudo-views. Lastly, hybrid data through staged Gaussian primitives optimization, balancing rendering quality and generalization. Simulation experiments highlight the importance of the generalizability metric, SDP, while comparisons on ScanNet++ and custom datasets demonstrate our approach's superiority over previous work in handling challenging regions.

\textbf{Limitations and Future Works.} The proposed depends on the quality of the pseudo-views; however, when the field of pseudo-view is narrow, the model struggles to correlate image content with the source views, leading to geometric ambiguity at the edges. Therefore, we will consider region-based training to reduce restoration uncertainty.

\section*{Acknowledgment}
This work was supported in part by the National Natural Science Foundation of China under Grant U24A20249,  in part by the Key R\&D Program of Zhejiang 2025C01075, and in part by the Key R\&D Program of Ningbo under Grant 2023Z060, 2023Z220.

\bibliographystyle{IEEEtran}
\bibliography{ref}

\end{document}